\documentclass[sigconf]{acmart}
\usepackage{multirow}
\AtBeginDocument{%
	\providecommand\BibTeX{{%
			\normalfont B\kern-0.5em{\scshape i\kern-0.25em b}\kern-0.8em\TeX}}}

\copyrightyear{2022}
\acmYear{2022}
\setcopyright{acmcopyright}\acmConference[MM '22]{Proceedings of the 30th ACM
	International Conference on Multimedia}{October 10--14, 2022}{Lisboa, Portugal}
\acmBooktitle{Proceedings of the 30th ACM International Conference on Multimedia
	(MM '22), October 10--14, 2022, Lisboa, Portugal}
\acmPrice{15.00}
\acmDOI{10.1145/3503161.3547952}
\acmISBN{978-1-4503-9203-7/22/10}

\acmSubmissionID{826}

\begin{document}

\title{Enhancement by Your Aesthetic: An Intelligible Unsupervised Personalized Enhancer for Low-Light Images}

%
\author{Naishan Zheng}
\authornotemark[1]
\affiliation{
	  \institution{University of Science and Technology of China}
	  \city{}
	  \state{}
	  \country{}
	  \postcode{}
	}
\email{nszheng@mail.ustc.edu.cn}

\author{Jie Huang}
\authornote{Both authors contributed equally to this research.}
\affiliation{
	  \institution{University of Science and Technology of China}
		\city{}
		\state{}
		\country{}
		\postcode{}
}
\email{hj0117@mail.ustc.edu.cn}

\author{Qi Zhu}
\affiliation{%
	  \institution{University of Science and Technology of China}
		\city{}
		\state{}
		\country{}
		\postcode{}
}
\email{zqcrafts@mail.ustc.edu.cn}

\author{Man Zhou}
\affiliation{%
	  \institution{University of Science and Technology of China}
		\city{}
		\state{}
		\country{}
		\postcode{}
	}
\email{manman@mail.ustc.edu.cn}

\author{Feng Zhao}
\authornote{Feng Zhao is the corresponding author.}
\affiliation{%
	  \institution{University of Science and Technology of China}
		\city{}
		\state{}
		\country{}
		\postcode{}
	}
\email{fzhao956@ustc.edu.cn}

\author{Zheng-Jun Zha}
\affiliation{%
	  \institution{University of Science and Technology of China}
		\city{}
		\state{}
		\country{}
		\postcode{}
}
\email{zhazj@ustc.edu.cn}

%


\begin{abstract}
Low-light image enhancement is an inherently subjective process whose targets vary with the user’s aesthetic. 
Motivated by this, several personalized enhancement methods have been investigated. 
However, the enhancement process based on user preferences in these techniques is invisible, \emph{i.e.}, a "black box". 
In this work, we propose an intelligible unsupervised personalized enhancer (iUP-Enhancer) for low-light images, which establishes the correlations between the low-light and the unpaired reference images with regard to three user-friendly attributions (brightness, chromaticity, and noise). 
The proposed iUP-Enhancer is trained with the guidance of these correlations and the corresponding unsupervised loss functions. 
Rather than a “black box” process, our iUP-Enhancer presents an intelligible enhancement process with the above attributions. 
Extensive experiments demonstrate that the proposed algorithm produces competitive qualitative and quantitative results while maintaining excellent flexibility and scalability. 
This can be validated by personalization with single/multiple references, cross-attribution references, or merely adjusting parameters.

\end{abstract}





\maketitle

\begin{figure}[!t]
	\centering
	\includegraphics[width=0.32\textwidth]{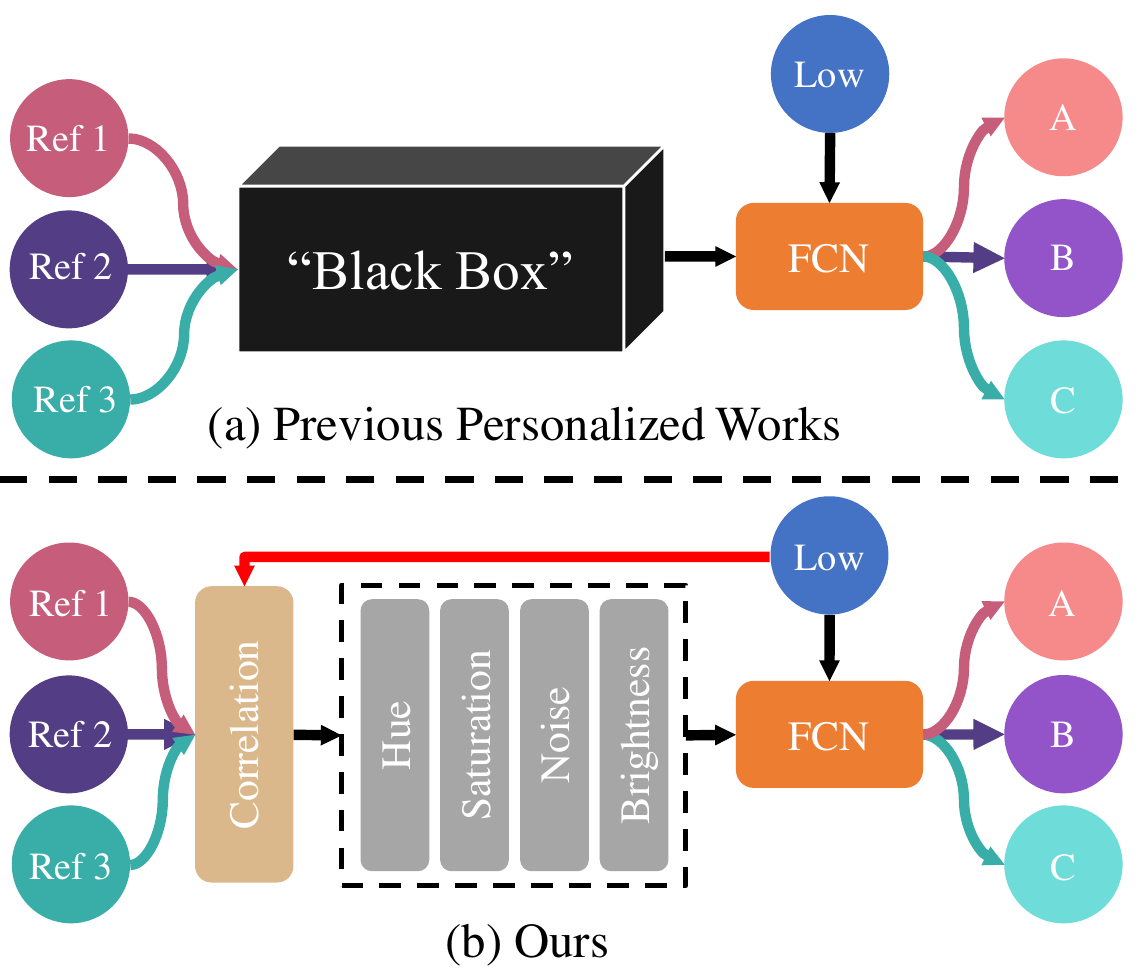}
	\vspace{-0.45cm}
	\caption{Comparison between previous works and our proposed method. (a) Previous works extract features from the reference images by stacked convolutional layers, resulting in an invisible enhancement process, just like a "black box"; and (b) our iUP-Enhancer provides an intelligible solution by calculating the brightness, chromaticity, and noise-level correlations between the low-light and reference images.}
	\vspace{-0.5cm}
	\label{fig:tissue}
\end{figure}

\section{Introduction}
The widespread of smartphone cameras has reduced the obstacles to photography dramatically. 
However, images captured under low-light conditions exhibit undesirable characteristics, such as low contrast and implicit noise. 
To this end, various deep-learning-based low-light image enhancement (LLIE) algorithms have been proposed to mitigate this effect by learning the mapping from low-light to normal-light. 
Undeniably, they have made great progress toward producing aesthetically pleasing images. 
However, these algorithms implicitly presume merely one deterministic output for a low-light image. 
This ill-posed assumption contradicts the consensus that different users have distinct aesthetic tastes. 
Thus, it is necessary to develop a personalized algorithm that yields customized results according to user preferences.

Recently, several approaches have been proposed for personalized enhancement, including
PieNet~\cite{PieNet}, StarEnhancer~\cite{starenhancer}, and cGAN~\cite{IJCAI}.
These methods convey the users' preferences through a collection of low-light images that they find fascinating. 
PieNet and StarEnhancer extract a style embedding from the reference images through a pre-trained style encoder, and then achieve personalized enhancement under the guidance of the extracted embedding. 
However, they require an extensive collection of reference images containing diverse styles due to the supervised strategy, limiting their generalization on the real-world scenarios. 
To handle this circumstance, cGAN jointly implements style embedding learning and customized enhancement in an unsupervised framework, which introduces the GAN to discriminate the embeddings of the enhanced and reference images. 
Unfortunately, the style embedding extracted by the stacked convolutional layers hides the enhancement process, \emph{i.e.}, a "black box" [see Fig.~\ref{fig:tissue}(a)]. 
In addition, ReLLIE~\cite{ReLLIE} enables the users to interactively specify the number of enhancement steps. 
Regrettably, it is inadequate to adjust brightness without altering the style. 
In summary, existing personalized  enhancement techniques have the following shortcomings: 
(1) collecting enormous reference images with multiple styles is time-consuming, limiting the widespread application in reality; and 
(2) the "black box" makes the personalized enhancement process invisible, resulting in poor flexibility for users' customization.

To address the above issues, we design an intelligible unsupervised personalized enhancer for low-light images, namely iUP-Enhancer. 
The proposed iUP-Enhancer is implemented in an unsupervised manner by establishing the correlations between the input and the unpaired reference images, and applying the constraints imposed by the unsupervised loss functions. 
To achieve an intelligible enhancement process, we derive the correlations of three user-friendly attributions, including brightness, chromaticity, and noise level. 
The enhanced image will then be customized under the guidance of these correlations, as described in Fig.~\ref{fig:framework}. 
Specifically, based on the Retinex theory, the low-light and reference images can be decomposed into corresponding illumination and reflectance maps. 
As illustrated in Fig.~\ref{fig:detail}, the brightness adjustment is accomplished by pulling the intensity distribution of the illumination maps for the reference and low-light images. 
The chromaticity is modulated by the hue and saturation correlations in the HSV space of their reflectance maps. 
The denoising effect is guided by the correlation of the noise-level maps on the reflectance components. 
With the guidance of these correlations learned from the unpaired images, our iUP-Enhancer presents an intelligible enhancement process for personalization in an unsupervised manner. 

In contrast to the enhancement process hidden in a "black box", our iUP-Enhancer presents users with intelligible enhancement operations, including brightness, chromaticity, and noise level. 
During the inference, users are enabled to adjust particular attributions for personalization by single/multiple references, cross-attribution references, or adjusting parameters. 
Experiments demonstrate the excellent flexibility and scalability of the proposed framework for producing plausible results according to users' preferences. 

Our contributions are summarized as follows:
\begin{itemize}
	\item We propose an intelligible unsupervised personalized enhancer for low-light images, namely iUP-Enhancer, which establishes the correlations of multiple attributions between the input and the unpaired reference images.
	\item The iUP-Enhancer is trained with the guidance of correlations in brightness, chromaticity, and noise level, and is constrained by the corresponding unsupervised loss functions, thus presenting an intelligible enhancement process.
	\item Experimental results demonstrate that our iUP-Enhancer produces competitive qualitative and quantitative results. Moreover, it embraces flexibility and scalability by performing personalization with single/multiple references, cross-attribution references, or merely adjusting parameters.
\end{itemize}

\section{Related Work}
\subsection{Conventional Methods}
The earliest low-light image enhancement techniques can be classified into two main categories: histogram equalization (HE) and gamma correction. 
However, these methods exhibit poor generalization ability when being applied to diverse real-world low-light images.
In addition, several methods have been proposed based on the Retinex~\cite{land1977retinex} theory that decomposes a color image into two components: illumination and reflectance. 
Guo \emph{et al.}~\cite{LIME} estimated the maximum intensity of RGB channels as the initial coarse illumination map and refined it based on a structure prior. 
Li \emph{et al.}~\cite{Roubust_retinex} proposed a robust Retinex model that estimates an extra noise map.

\begin{figure}[t]
	\centering
	\includegraphics[width=0.4\textwidth]{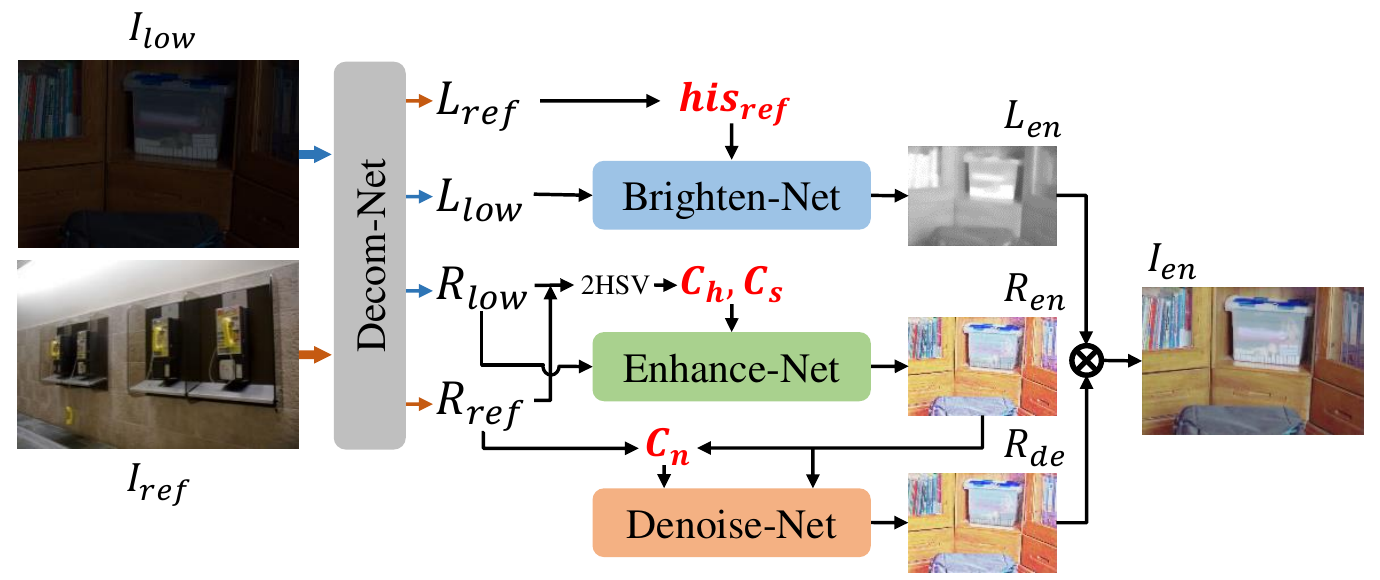}
	\vspace{-0.3cm}
	\caption{The overview of the proposed iUP-Enhancer. The low-light and reference images are decomposed into the illumination and reflectance components based on the Retinex theory. The illumination map is customized by the brightness histogram of the reference $(his_{ref})$, while the reflectance map is enhanced through the correlation on chromaticity $([C_{h}, C_{s}])$ and denoised by the correlation on noise level $(C_{n})$. They are multiplied to generate the final enhanced result.}
	\label{fig:framework}
	\vspace{-0.5cm}
\end{figure}

\begin{figure*}[!t]
	\centering
	\includegraphics[width=0.92\textwidth]{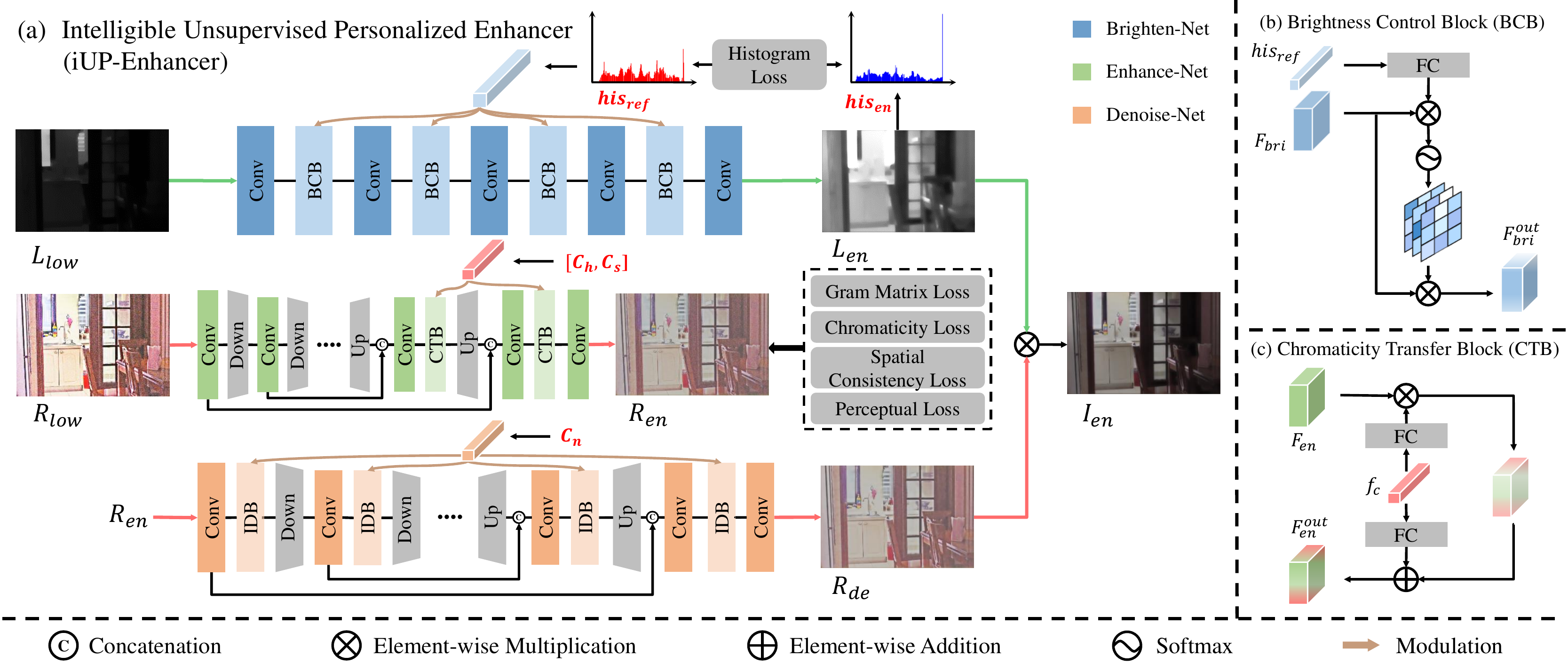}
	\vspace{-0.4cm}
	\caption{The details of our iUP-Enhancer. (a) The iUP-Enhancer presents an understandable enhancement process consisting of brightness, chromaticity, and noise level. The Brighten-Net takes $his_{ref}$ as the input to customize the illumination map, while the correlations on chromaticity and noise level, $[C_{h}, C_{s}]$, and $C_{n}$, are utilized to adjust the reflectance map by the Enhance-Net and the Denoise-Net, respectively. The correlations are utilized to guide the enhancement process by (b) the brightness control block (BCB), (c) the chromaticity transfer block (CTB), and the interactive denoising block (same as the BCB), respectively.}
	\vspace{-0.1cm}
\label{fig:detail}

\end{figure*}

\subsection{Learning-based Methods}
Recently, deep learning-based algorithms~\cite{LLNet,KinD,progressive_retinex,LPNet,DSLR,ExCNet,faset_uniform_enh,semantic_retinex,RetinexDIP,yao2019spectral,pan2022towards,wang2021jpeg} have been proposed to improve low light and other degradations. 
Zhao \emph{et al.}~\cite{INN_enhance} employed the invertible neural network to perform bidirectional feature learning between the low-light and normal-light images. 
LLFlow~\cite{LLFlow} proposes to model the well-exposed image as a conditional distribution and samples from it to improve the underexposed image, achieving the one-to-many prediction.
However, these supervised algorithms require paired training data that are not applicable to the real-world scenarios. 
To eliminate the dependence of paired data, unsupervised strategies have been introduced to the LLIE.  
ZERO-DCE~\cite{ZERO_DCE} predicts a pixel-wise mapping to adjust the dynamic range of the low-light images without any paired or unpaired data. 
RUAS~\cite{RUAS} develops a non-reference learning strategy to discover low-light priors and unroll their optimization processes from a compact search space. 
However, the above solutions presume that a low-light image has a single deterministic output, revealing their shortcoming that the prediction cannot vary with diverse preferences of each individual.

\subsection{Personalization}
ReLLIE~\cite{ReLLIE} applies reinforcement learning to produce pixel-wise curves, where users for personalization can customize the number of enhancement steps.
PieNet~\cite{PieNet} and StarEnhancer~\cite{starenhancer} prefetch the features from the reference images into a style embedding, which customizes the enhancer via adaptive instance normalization. 
cGAN~\cite{IJCAI} formulates the personalized LLIE as a modulation code learning task by discriminating the style embedding of the reference images and the enhanced images. 
However, the style embedding extracted by the stacked convolutional layers hides the enhancement process, \emph{i.e.}, a "black box".

\section{Methods}

\subsection{Overview}
To enhance the input low-light image $I_{low}$ to the normal-light one $I_{en}$ while varying with the reference image $I_{ref}$ users provided, we propose an intelligible unsupervised personalized enhancer, termed as iUP-Enhancer. 
Fig.~\ref{fig:framework} describes the structure of our iUP-Enhancer. 
Firstly, based on the Retinex theory, $I_{low}$ and unpaired $I_{ref}$ are decomposed into the corresponding illumination and reflectance maps by the Decom-Net, \emph{i.e.}, $L_{low}$ and $R_{low}$, $L_{ref}$ and $R_{ref}$. 

To provide users with an understandable enhancement process, we extract the correlations between the components in terms of brightness, chromaticity, and noise level to guide the personalized enhancement process. 
Specifically, (1) brightness: the Brighten-Net takes $L_{low}$ and the brightness histogram of $L_{ref}$ ( $his_{L_{ref}}$) as the input to predict the enhanced illumination map $L_{en}$. 
(2) Chromaticity: the chromaticity correlation comprises hue correlation $C_{h}$ and saturation correlation $C_{s}$, which are derived by converting $R_{low}$ and $R_{ref}$ into the HSV space and then calculating their hue and saturation channel correlations. 
The Enhance-Net generates the enhanced reflectance map $R_{en}$ with the assist of $C_{h}$ and $C_{s}$, which matches the chromaticity of $R_{ref}$. 
(3) Noise level: the Denoise-Net performs particular denoising effect based on the noise-level correlation $C_n$ between $R_{en}$ and $R_{ref}$ to produce the denoised reflectance map $R_{de}$. 
We integrate $L_{en}$ and $R_{de}$ as the personalized enhanced result $I_{en}$. 
The above process is implemented in an unsupervised manner by leveraging these correlations and the corresponding loss functions.

\subsection{Personalized Enhancement}
This section details how to perform an intelligible personalized enhancement process with three attributions, including brightness, chromaticity, and noise level.

\textbf{Decomposition.}
The Decom-Net learns to decompose the input image into the illumination and reflectance components under the prior that the low-light and normal-light images share an identical reflectance map. 
However, the corresponding ground truth is unavailable. 
Thus, we follow the setting of the architecture and the loss functions in RRDNet~\cite{RRDNet} to train our Decom-Net. 
More details will be provided in the supplement. 
In the follow-up training, the pre-trained Decom-Net decomposes $I_{low}$ and $I_{ref}$ into $L_{low}$ and $R_{low}$, $L_{ref}$ and $R_{ref}$, respectively. 
\vspace{0.2cm}

\textbf{Brightness.}
Following the decomposition, a Brighten-Net [see Fig. ~\ref{fig:detail}(a)] is designed to brighten the illumination map $L_{low}$. 
To bring the brightness adjustment ability to the network with $L_{ref}$ adaptively, we introduce a brightness control block (BCB)~\cite{SFT}. 
As described in Fig.~\ref{fig:detail}(b), it modulates the input features $F_{bri}$ with the guidance of the attention map derived by $his_{L_{ref}}$ as:

\begin{equation}
F_{bri}^{out} = F_{bri} \ast softmax(F_{bri} \odot v(his_{L_{ref}})), 
\end{equation}
where $v(\cdot)$ is the coefficient mapping achieved by two fully-connected (FC) layers, and $\odot$ denotes the channel-wise multiplication. 

\emph{Histogram Loss.} To derive $his_{ref}$, we introduce the histogram loss $L_{his}$ to minimize the distribution between $L_{en}$ and $L_{ref}$:

\vspace{-0.1cm}
\begin{equation}
L_{his}=\sum_{i}^{N} \left| his_{L_{ref}}^{i} - his_{L_{en}}^{i} \right|,
\end{equation}
where $his_{L_{en}}$ is the histogram of $L_{en}$, $N (=256)$ is the number of bins in the histogram, and $|\cdot|$ means the absolute value operation.

\begin{figure}[t]
\centering
\includegraphics[width=0.42\textwidth]{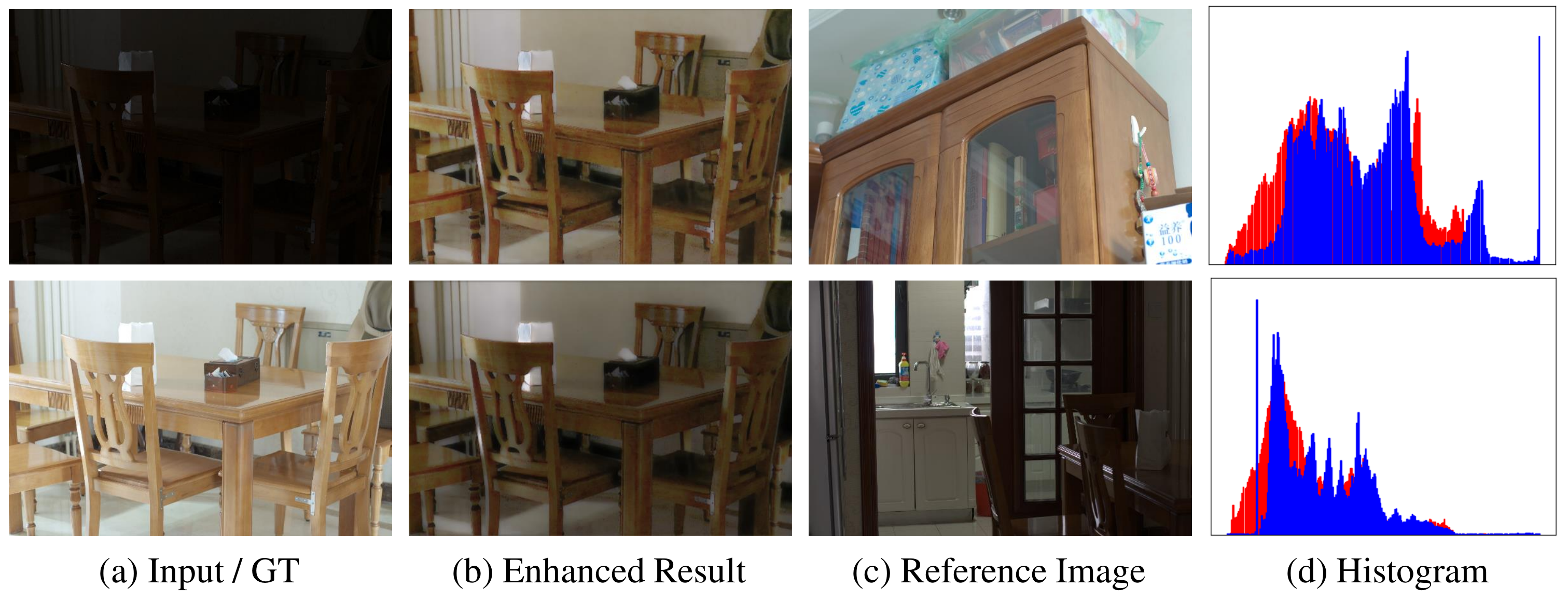}
\vspace{-0.3cm}
\caption{The brightness histogram of the personalized results guided by reference images. (a) The input low-light image $I_{low}$ and the ground truth, (b) the personalized enhanced results $I_{en}$ based on (c) the reference images $I_{ref}$ and (d) the brightness histogram of the reference images (blue) and the enhanced results (red).}
\label{fig:his}
\vspace{-0.4cm}
\end{figure}

As demonstrated in Fig.~\ref{fig:his}, with the above design, our iUP-Enhancer generates enhanced personalized results with a brightness distribution that approximates that of the reference image.
\vspace{0.2cm}

\textbf{Chromaticity.}
The Enhance-Net as shown in Fig.~\ref{fig:detail}(a), is proposed to transfer the chromaticity of $R_{ref}$ to $R_{en}$, since the chromaticity significantly reflects the aesthetics of users. 
Exactly, the HSV color space includes three channels, \emph{i.e.}, hue (H), saturation (S), and value (V), where the first two comprise the chromaticity. 
Consequently, we convert $R_{low}$ and $R_{ref}$ into the HSV space and obtain their histogram  $his_{R_{low}}^{h}$ and $his_{R_{ref}}^{h}$, $his_{R_{low}}^{s}$ and $his_{R_{ref}}^{s}$.
The correlations $C_{h}$ and $C_{s}$ between them reflect the differences in chromaticity: 

\vspace{-0.2cm}
\begin{equation}
C_{h} = \frac{his_{R_{low}}^{h} \cdot his_{R_{ref}}^{h}}{\|his_{R_{low}}^{h}\|_{2}\|his_{R_{ref}}^{h}\|_{2}},
C_{s} = \frac{his_{R_{low}}^{s} \cdot his_{R_{ref}}^{s}}{\|his_{R_{low}}^{s}\|_{2}\|his_{R_{ref}}^{s}\|_{2}},
\end{equation}
where $\cdot$ means dot product, and $\| \cdot \|_{2}$ denotes the $L_{2}$ normalization.

To transfer the chromaticity with interactivity, we implement the Enhance-Net with a chromaticity transfer block (CTB)~\cite{huang2017arbitrary}. 
As shown in Fig.~\ref{fig:detail}(c), we first obtain the chromaticity codes $f_{c}$ by fetching  $[C_{h}, C_{s}]$ through FC layers, and then the CTB modulates the features $F_{en}$ according to $f_{c}$:

\vspace{-0.1cm}
\begin{equation}
F_{en}^{out}=\gamma \odot (\frac{F_{en} - \mu }{\sigma } ) + \beta,
\end{equation}
where $\mu$ and $\sigma$ are the channel-wise mean and variance of $F_{en}$, and $\gamma$ and $\beta$ are the affine parameters learning from $f_{c}$ through FC layers.

To constrain $R_{en}$, we adopt the following unsupervised losses:

\emph{Gram Matrix Loss.} We apply the Gram matrix loss~\cite{gram} to measure the style difference between $R_{ref}$ and $R_{en}$:

\vspace{-0.1cm}
\begin{equation}
L_{gram}=\left | R_{ref} \cdot R_{ref}^{T} - R_{en} \cdot R_{en}^{T} \right |,
\end{equation}
where $T$ stands for the matrix transpose operation.

\emph{Chromaticity Loss.} The chromaticity loss encourages to pull the distance between the chromaticity distributions of $R_{low}$ and $R_{en}$: 

\vspace{-0.1cm}
\begin{equation}
L_{chr}=\sum_{i}^{N} \left| his_{R_{ref}}^{h_{i}} - his_{R_{en}}^{h_{i}} \right| + \sum_{i}^{N} \left| his_{R_{ref}}^{s_{i}} - his_{R_{en}}^{s_{i}} \right|,
\end{equation}
where $his_{R_{en}}^{h}$ and $his_{R_{en}}^{s}$ are hue and saturation histograms of $R_{en}$.

\begin{figure}[t]
	\centering
	\includegraphics[width=0.42\textwidth]{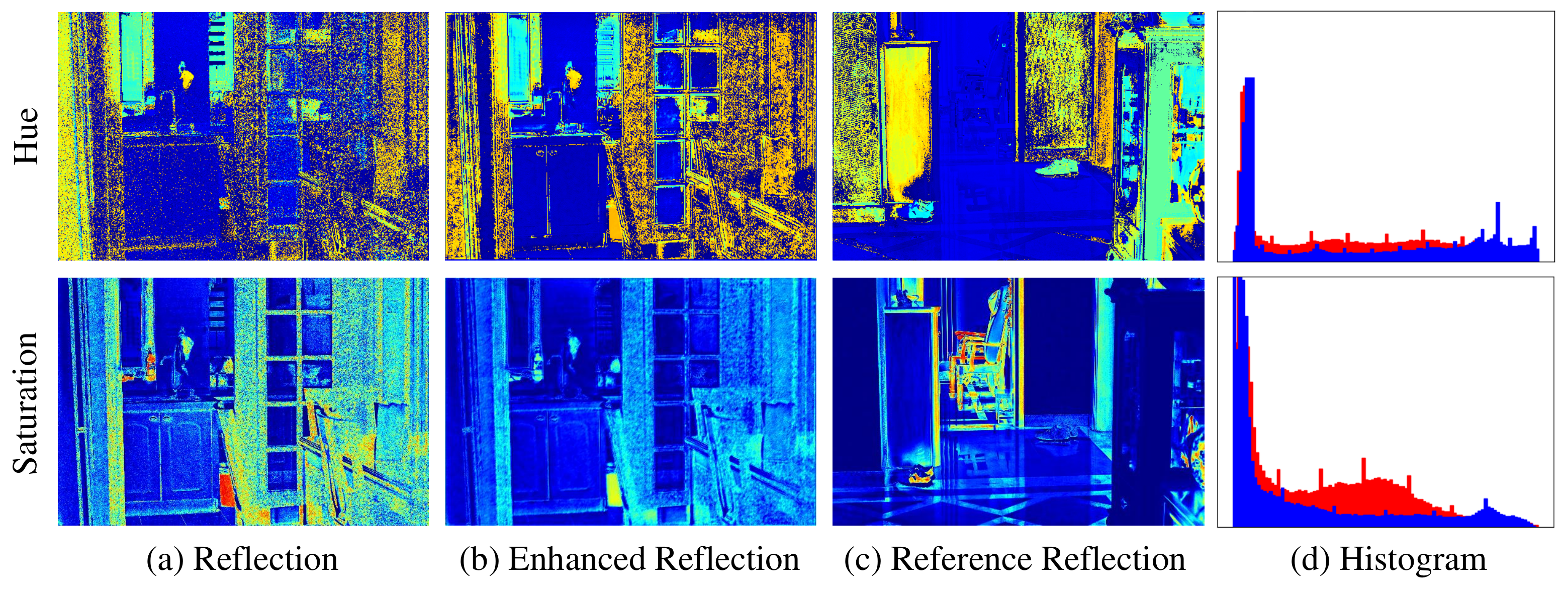}
	\vspace{-0.3cm}
	\caption{The first to third columns separately visualize the hue and saturation channel of the reflectance $R_{low}$, the enhanced reflectance $R_{en}$, and the reference reflectance $R_{ref}$. The fourth column shows the histogram of the reference reflectance (blue) and the enhanced reflectance (red). }
	\label{fig:hsv}
	\vspace{-0.4cm}
\end{figure}

\begin{figure*}[t]
\centering
\includegraphics[width=0.85\textwidth]{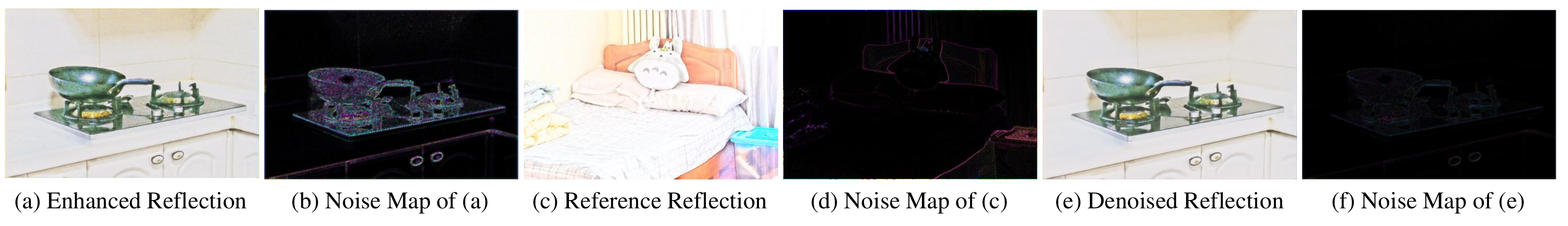}
\vspace{-0.4cm}
\caption{The noise maps of (a) the enhanced reflectance $R_{en}$, (c) the reference reflectance $R_{ref}$, and (e) the denoised reflectance $R_{de}$. The denoising effect of (e) is determined by the correlation between (b) and (d).}
\label{fig:denoise}
\vspace{-0.2cm}
\end{figure*}

\emph{Spatial Consistency Loss.} The spatial consistency loss evaluates the difference of neighboring regions between $R_{low}$ and $R_{en}$  on the horizontal and vertical gradients, which is defined as:

\vspace{-0.1cm}
\begin{equation}
L_{spa}=\| \nabla R_{low}^{4} - \nabla R_{en}^{4} \|^{2}_{2},
\end{equation}
where $ R_{low}^{4}$ and $R_{en}^{4}$ separately denote the $4 \times 4$ average pooling results of $R_{low}$ and $R_{en}$, and $\nabla$ represents the horizontal ($\nabla_{x}$) and vertical ($\nabla_{y}$) gradient operations.

\emph{Perceptual Loss.} To preserve the content information, we employ a perceptual loss as:

\vspace{-0.1cm}
\begin{equation}
L_{per}=\|\phi\left(R_{low}\right) - \phi\left(R_{en}\right)\|^{2}_{2},
\end{equation}
where $\phi\left(R_{low}\right)$ and $\phi\left(R_{en}\right)$ represent the features of $R_{low}$ and $R_{ref}$, respectively. Here, $\phi$ denotes the fifth-layer feature map of the pre-trained VGG19~\cite{VGG} on the ImageNet~\cite{imagenet}.

The fourth column in Fig.~\ref{fig:hsv} presents the histograms of $R_{en}$ (red) and $R_{ref}$ (blue) of the hue and saturation channels, which proves their similarity in chromaticity.
\vspace{0.2cm}

\noindent \textbf{Noise Level.} 
Another essential point is removing the noise on the reflectance map, thus we design a Denoise-Net [see Fig.~\ref{fig:detail}] for implementation. 
To evaluate the noise level and perform the personalized denoising effect, we follow~\cite{LLFlow} to estimate the noise maps. 
As shown in Fig.~\ref{fig:denoise}, the noise-level map of $R_{ref}$, $N_{ref}$, is lower than that of $R_{en}$, $N_{en}$. 
Meanwhile, different reference images have distinct noise levels, necessitating the particular denoising effect based on the correlation between $N_{en}$ and $N_{ref}$:

\begin{equation}
C_{n} = \frac{his_{N_{en}} \cdot his_{N_{ref}}}{\|his_{N_{en}}\|_{2}\|his_{N_{ref}}\|_{2}},
\end{equation}
where $his_{N_{ref}}$ and $his_{N_{en}}$ are the histograms of $N_{ref}$ and $N_{en}$.

To achieve adjustable denoising with the guidance of $C_{n}$, we pre-train a Denoise-Net in two steps (see Fig.~\ref{fig:denoiser}) that can customize the denoising effect via an interactive denoising block (IDB). 
In the first step, the network with $C_{n}=0$ is trained on the noisy images in the self-supervised manner~\cite{noise2noise} to obtain a stronger denoiser:

\begin{equation}
x = y + N(0, \sigma^{2}), \tilde{y} = Net(x, 0),
\end{equation}
where $x$ and $y$ are the noisy and clean images, $N(0, \sigma^{2})$ represents the Gaussian
distribution with a mean of 0 and a variance of $\sigma^{2}$, and $\tilde{y}$ denotes the denoised result. 
In the second step, the blended $\tilde{x}$ combines $\tilde{y}$ and $x$ with a weighting coefficient $\alpha$. 
We fix the backbone of the network and finetune the IDBs with $C_{n} = 1$: 

\begin{figure}[t]
\centering
\includegraphics[width=0.4\textwidth]{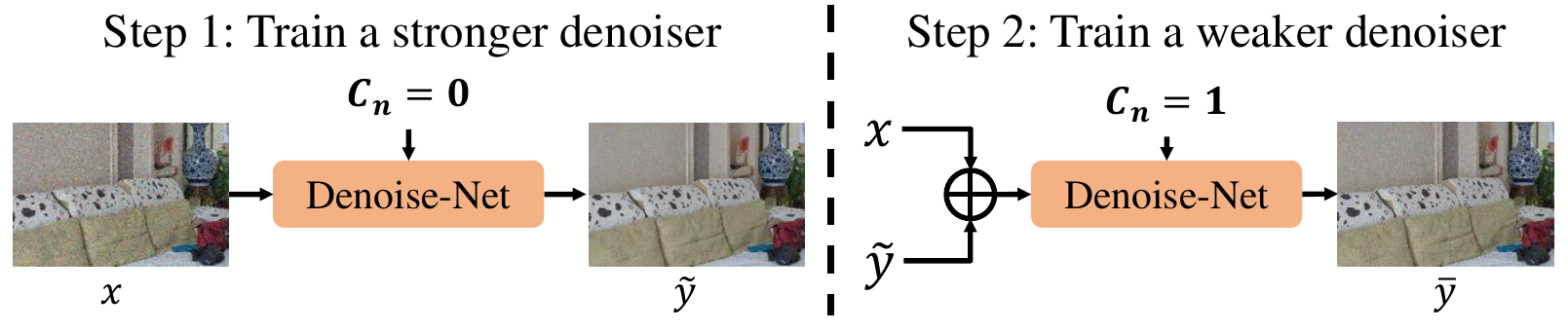}
\vspace{-0.4cm}
\caption{Training steps of the Denoise-Net.}
\label{fig:denoiser}
\vspace{-0.5cm}
\end{figure}

\begin{equation}
\tilde{x} = \alpha x + (1 - \alpha) \tilde{y},  \bar{y} = Net(\tilde{x}, 1),
\end{equation}
where $\alpha$ is a random value in a range of $[0.7, 0.9]$. The noise variance of $\tilde{x}$ is $(\alpha \sigma)^{2}$, thus a weaker denoiser will be obtained. 
Note that the IDB is evolved from the BCB module by replacing the softmax with sigmoid to avoid all-zero when $C_{n} = 0$.

In this way, the Denoise-Net is capable of achieving a controllable denoising effect, as proved in the supplement. 
During the enhancement, the $C_{n}$ will be mapped from $[-1, 1]$ to $[0, 1]$ to control the denoising effect of $R_{en}$. 
As shown in Fig.~\ref{fig:denoise}, the noise level of $R_{de}$ has significantly been reduced and approximated to $N_{ref}$. 

The total loss of our iUP-Enhancer is expressed as:
\vspace{-0.1cm}
\begin{equation}
L = L_{his} + w_{1} * L_{gram} + w_{2} * L_{chr} + w_{3} * L_{spa} + w_{4} * L_{per}.
\label{loss}
\end{equation}

\subsection{User Interaction}
\begin{figure}[t]
\centering
\includegraphics[width=0.37\textwidth]{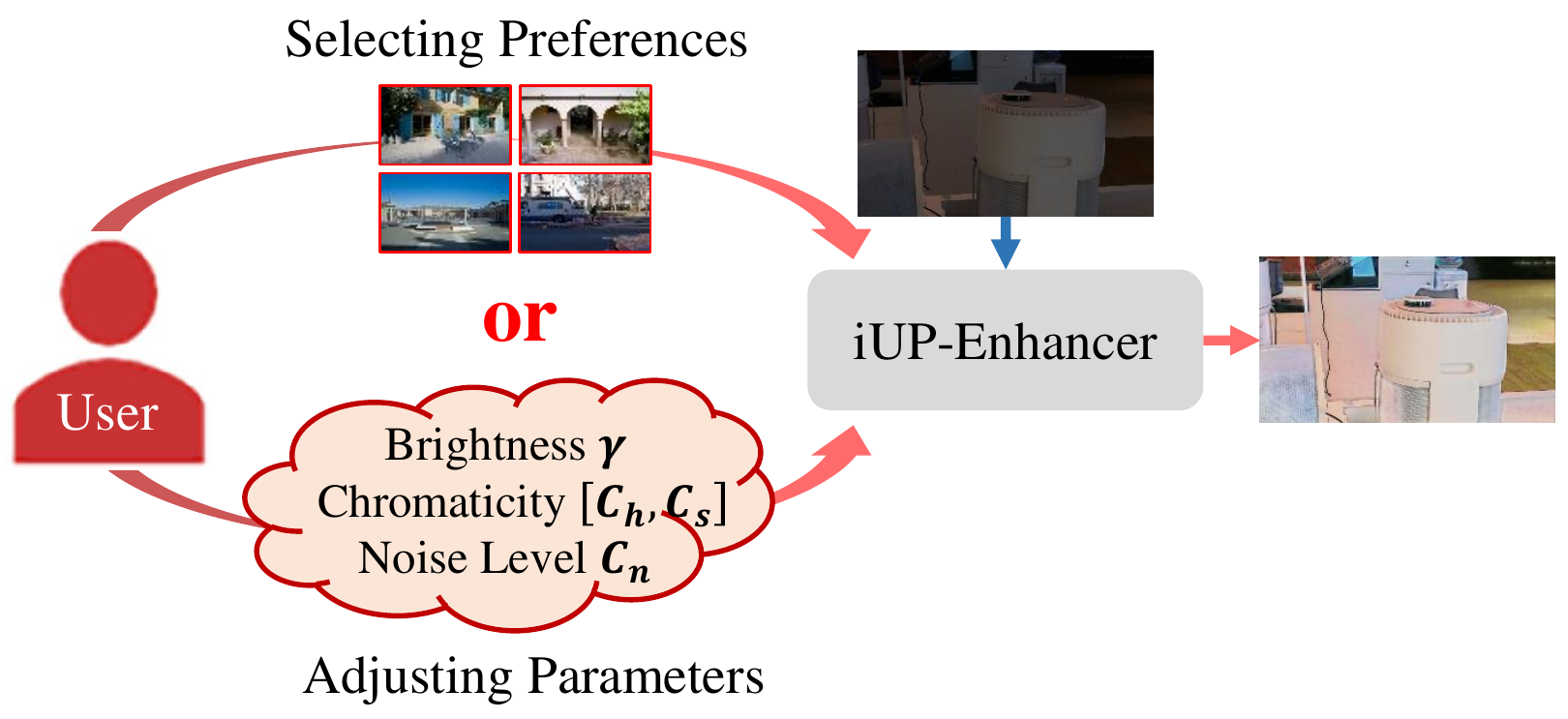}
\vspace{-0.4cm}
\caption{Approaches for users to customize enhanced images: (1) selecting preferred images or (2) adjusting parameters for brightness, chromaticity, and noise level.}
\label{fig:interaction}
\vspace{-0.4cm}
\end{figure}

Flexibility and scalability are the ability to accommodate a new user's aesthetic with minimal effort. 
Benefiting from the intelligible enhancement process presented by our iUP-Enhancer, including brightness, chromaticity, and noise level, we consider three simple-to-follow schemes for customization, as described in Fig.~\ref{fig:interaction}. 

\textbf{Single or multiple references. }
Users can select their preferred images to build correlations between the low-light image and their preferences. 
When there are multiple reference images, the correlations are determined by averaging the correlations of all. 
Then, with the guidance of these correlations, the pre-trained iUP-Enhancer yields personalized enhanced images (see Fig.~\ref{fig:per1}). 

\textbf{Cross-attribution references. }
The user may prefer the brightness of image A, the chromaticity of image B, and the noise level of image C, respectively. 
As illustrated in Fig.~\ref{fig:per3}, our iUP-Enhancer performs personalized enhancement by flexible cross-attribution references to meet this demand. 
In other words, three reference images consistent with the user's aesthetics are selected with regard to the above three aspects.

\textbf{Adjusting parameters. }
As shown in Fig.~\ref{fig:per2}, our iUP-Enhancer is scalable, allowing for personalized enhancement based on the brightness, chromaticity, and noise-level parameters specified by the users, \emph{i.e.}, $\gamma$, $[C_{h}, C_{s}]$, and $C_{n}$. 
The $\gamma$ denotes the coefficient in $L_{low}^{1/\gamma}$, whose brightness histogram, $his_{L_{low}}^{\gamma}$, is utilized to guide the brightness enhancement. 
This scheme simply requires the adjusting of parameters and does not need any reference images.

\begin{table*}[t]
	\caption{Quantitative comparison results of our iUP-Enhancer with state-of-the-art methods on the LOL and FiveK datasets. The reported PSNR and SSIM are performance averaged over five validations with randomly selected images from the reference set. US means the number of votes that each method obtains in the user studies. The best results are highlighted in bold.}
	\vspace{-0.3cm}
	\centering
	\setlength{\tabcolsep}{1.1mm}{
		\begin{tabular}{cc|cccccccccccc}
			\toprule
			\multirow{2}{*}{Dataset} & \multirow{2}{*}{Metrics} & \multicolumn{12}{c}{Method} \\
			\cline{3-14}
			&& CRM & Dong & JieP & LIME & SRIE & RRDNet & ZERO-DCE & RetinexDIP & RUAS & ReLLIE & cGAN & Ours \\
			\midrule
			\multirow{3}{*}{LOL} 
			& PSNR $\uparrow$
			& 14.5673 & 16.7165 & 9.1329 & 16.5029 & 12.3100 & 11.2279 & 15.2963 & 9.1581 & 16.4047 & 18.7362 & 15.1225 & \textbf{19.5893}\\
			&SSIM $\uparrow$
			& 0.6000 & 0.4721 & 0.3493 & 0.5663 & 0.5023 & 0.4472 & 0.5125 & 0.3387 & 0.5716 & 0.5609 & 0.6074 & \textbf{0.7245}\\
			&US $\uparrow$
			& 35 & 29 & 6 & 67 & 11 & 31 & 26 & 4 & 58 & 12 & 79 & \textbf{392}\\
			\midrule
			\midrule
			\multirow{2}{*}{FiveK} 
			& PSNR $\uparrow$
			& 14.8823 & 8.4639 & 15.2666 & 8.5613 & 15.9961 & 17.4061 & 12.3648 & 12.7014 & 9.5609 & 6.3932 & 16.8989 & \textbf{18.0260}\\
			& SSIM $\uparrow$
			& 0.6950 & 0.5457 & 0.7108 & 0.5916 & 0.5602 & 0.7112 & 0.6499 & 0.6849 & 0.5906 & 0.4715 & 0.7175 & \textbf{0.7262}\\
			\midrule 
			\midrule
			\multirow{2}{*}{Huawei} 
			& PSNR $\uparrow$
			& 13.6459 & 8.8148 & 14.0042 & 12.5659 & 13.6213 & 9.0527 & 13.0014 & 11.5465 & 14.5253 & 15.7388 & 14.7683 & \textbf{17.3729} \\
			& SSIM $\uparrow$
			& 0.5058 & 0.3423 & 0.5212 & 0.4966 & 0.5158 & 0.2138 & 0.4514 & 0.3850 & 0.5348 & 0.4954 & 0.4741 & \textbf{0.5771} \\
			\bottomrule
			\vspace{-0.7cm}
	\end{tabular}}
	\label{tab:LOLFiveK}
\end{table*}

\begin{figure*}[t]
\centering
\includegraphics[width=0.78\textwidth]{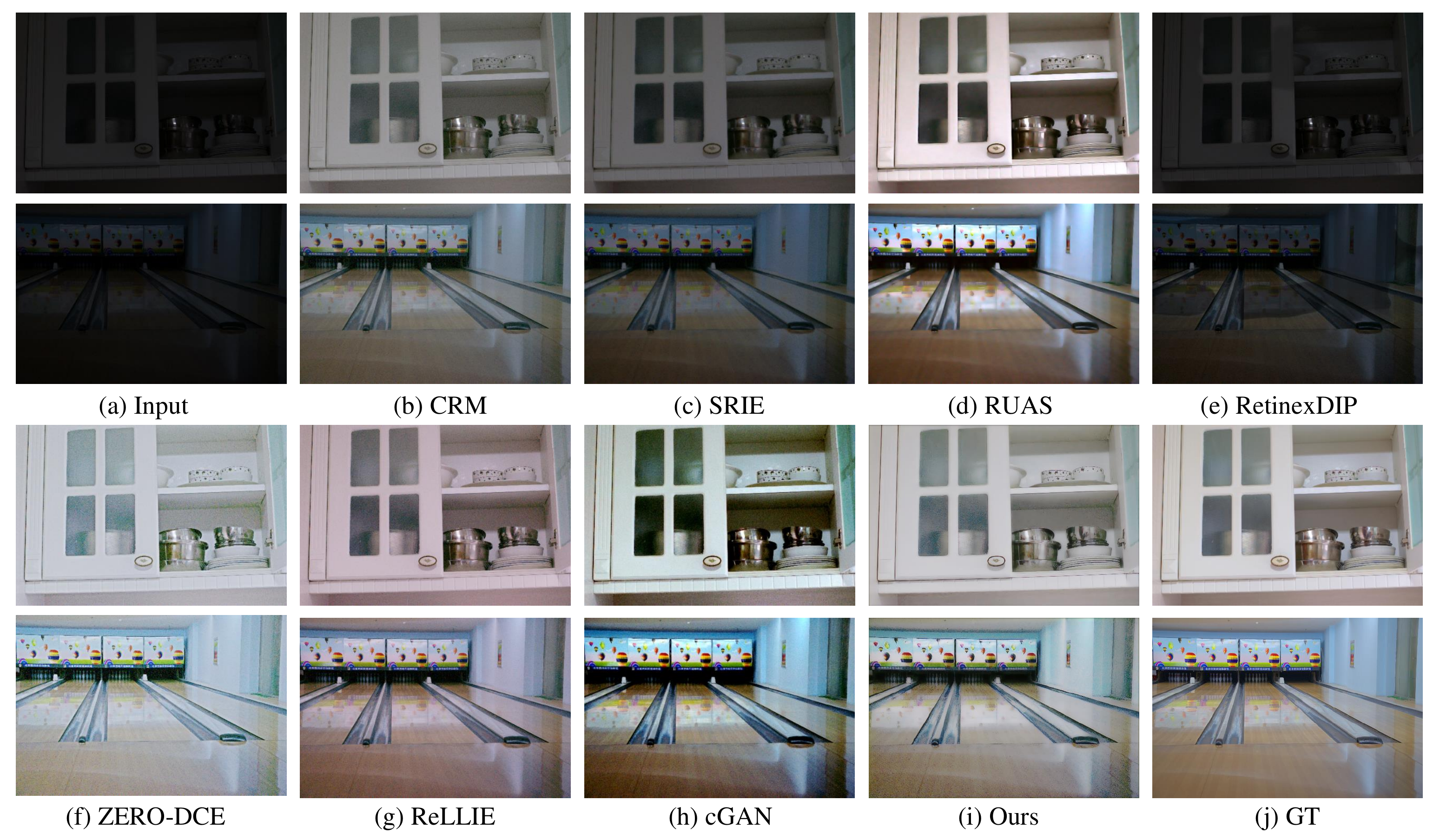}
\vspace{-0.4cm}
\caption{Visual comparison with state-of-the-art low-light image enhancement methods on the LOL dataset.}
\label{fig:LOL}
\vspace{-0.3cm}
\end{figure*}

\section{Experimental results}

\subsection{Experimental Settings}

\textbf{Datasets and baselines}
One of the primary advantages is that our network can be trained on unpaired low-light and normal-light reference images and learns various properties from an extensive collection of reference images. 
Thus, we collect reference images from three different datasets (LOL~\cite{KinD}, ExDark~\cite{ExDark}, and FiveK~\cite{FiveK}) and ignore the paired information. 
We conduct experiments on three standard datasets with paired data (LOL~\cite{KinD}, FiveK~\cite{FiveK}, and Huawei~\cite{hai2021r2rnet}). 
The training and testing sets of the LOL data in our experiments are identical to the official selection.
For the FiveK dataset, we randomly split 5000 images into 4750 images for training and 250 images for testing.
For the Huawei dataset, we randomly select 280 images for testing, and the remaining 2200 images for training. We compare our iUP-Enhancer with the state-of-the-art approaches including five conventional methods (CRM~\cite{CRM}, Dong~\cite{Dong}, JieP~\cite{JIEP}, LIME~\cite{LIME}, and SRIE~\cite{SRIE}), four unsupervised methods (RRDNet~\cite{RRDNet}, ZERO-DCE~\cite{ZERO_DCE}, RetinexDIP~\cite{RetinexDIP}, RUAS~\cite{RUAS}), and two personalized low-light enhancement methods (ReLLIE~\cite{ReLLIE}, cGAN~\cite{IJCAI}). 
For our method and the supervised techniques, the results on the paired datasets are retrained on them.

\textbf{Implementation details}
Our model is implemented in PyTorch, and the training is carried out on a single NVIDIA GTX 1080 TI. We adopt the Adam optimizer with $\beta_{1}=0.9$, $\beta_{2}=0.99$ for a total of $300K$ iterations. 
The initial learning rate is set to $1 \times 10^{-4}$ and decreases with a factor of $0.5$ every $50k$ iterations. 
A batch size of $18$ is applied. 
We fix the hyper-parameters in Eq.~\ref{loss}, \emph{i.e.}, $w_{1} = 0.0001$, $w_{2} = 0.01$, $w_{3} = 0.03$, and $w_{4} = 0.001$. 
The decomposition network is pre-trained on the LOL dataset. 
Similarly, the training is done using the ADAM optimizer, and the learning rate remains constant with $1 \times 10^{-4}$. The batch and patch sizes are set to $16$ and $48 \times 48$.

\begin{figure*}[t]
\centering
\includegraphics[width=0.78\textwidth]{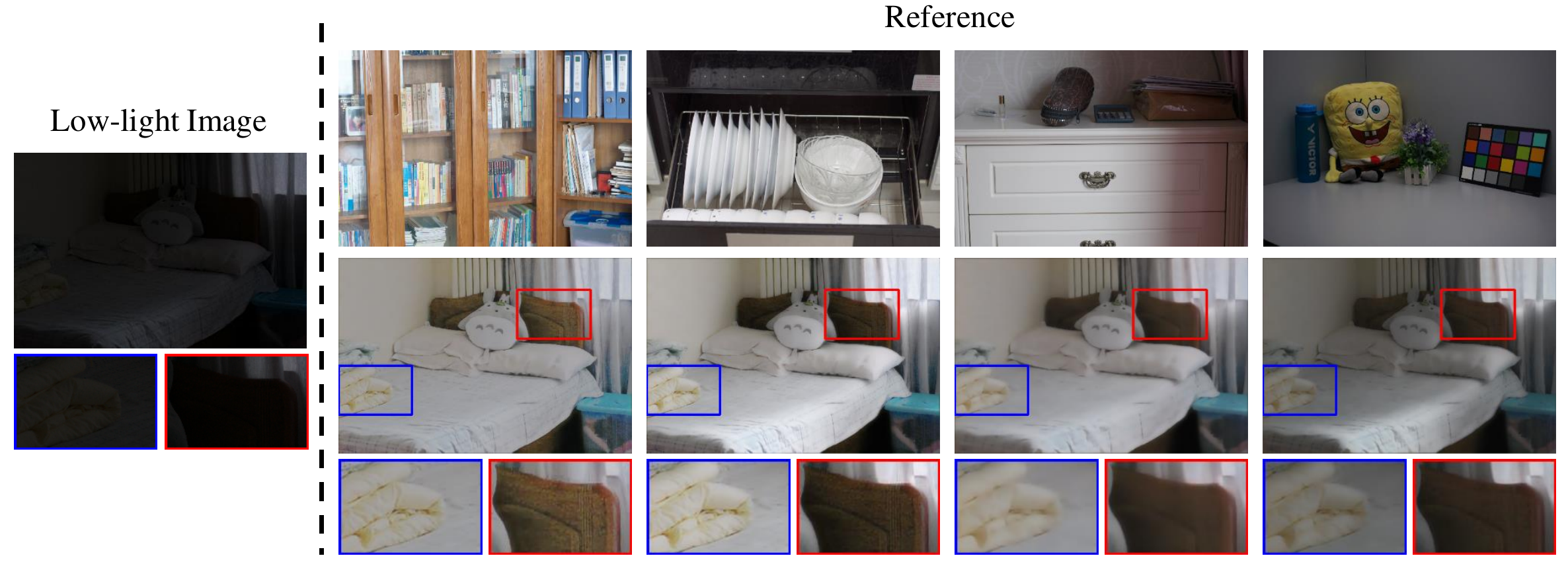}
\vspace{-0.4cm}
\caption{Examples of customized low-light image enhancement with single reference images.}
\vspace{-0.2cm}
\label{fig:per1}
\end{figure*}

\begin{figure*}[t]
\centering
\includegraphics[width=0.78\textwidth]{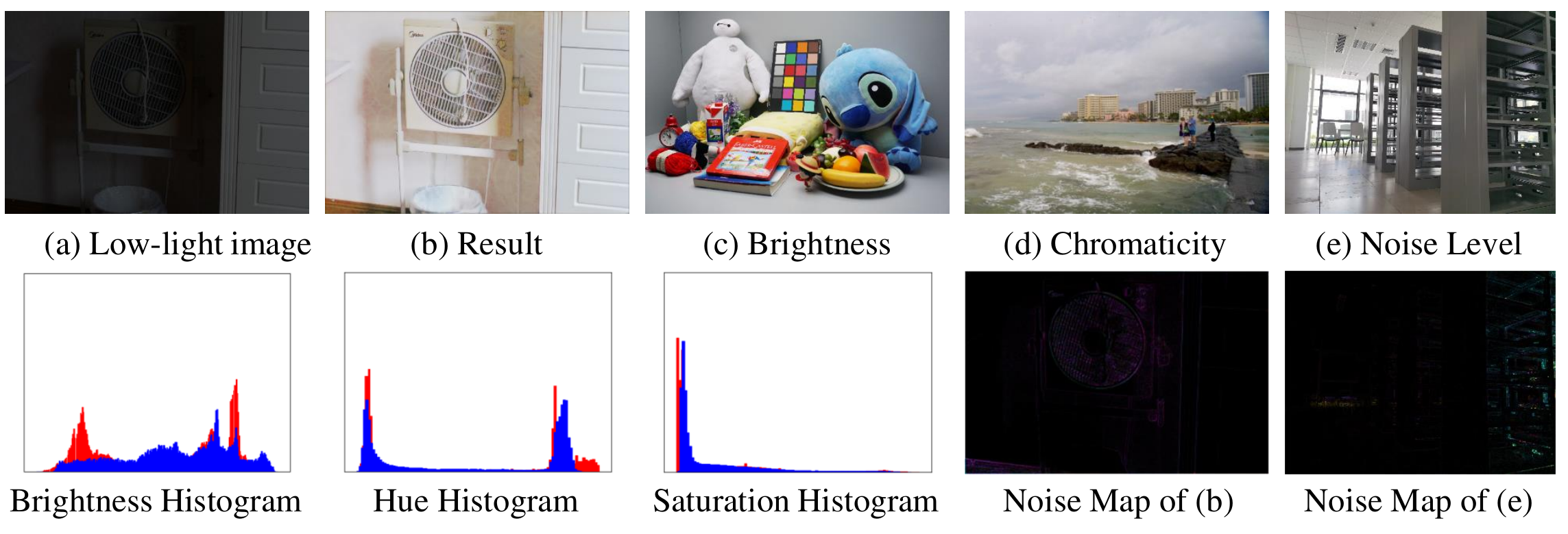}
\vspace{-0.5cm}
\caption{Examples of customized low-light image enhancement with cross-attribution references. (a) Input low-light image, (b) enhanced result, and (c)-(e) samples from LOL, FiveK, and Huawei datasets, corresponding to the reference images of the three attributes (brightness, chromaticity, and noise level). The second row presents the brightness, hue, and saturation histograms of the reference image (blue) and the enhanced result (red), and the noise maps of (b) and (e).}
\vspace{-0.2cm}
\label{fig:per3}
\end{figure*}

\begin{figure*}[t]
\centering
\includegraphics[width=0.78\textwidth]{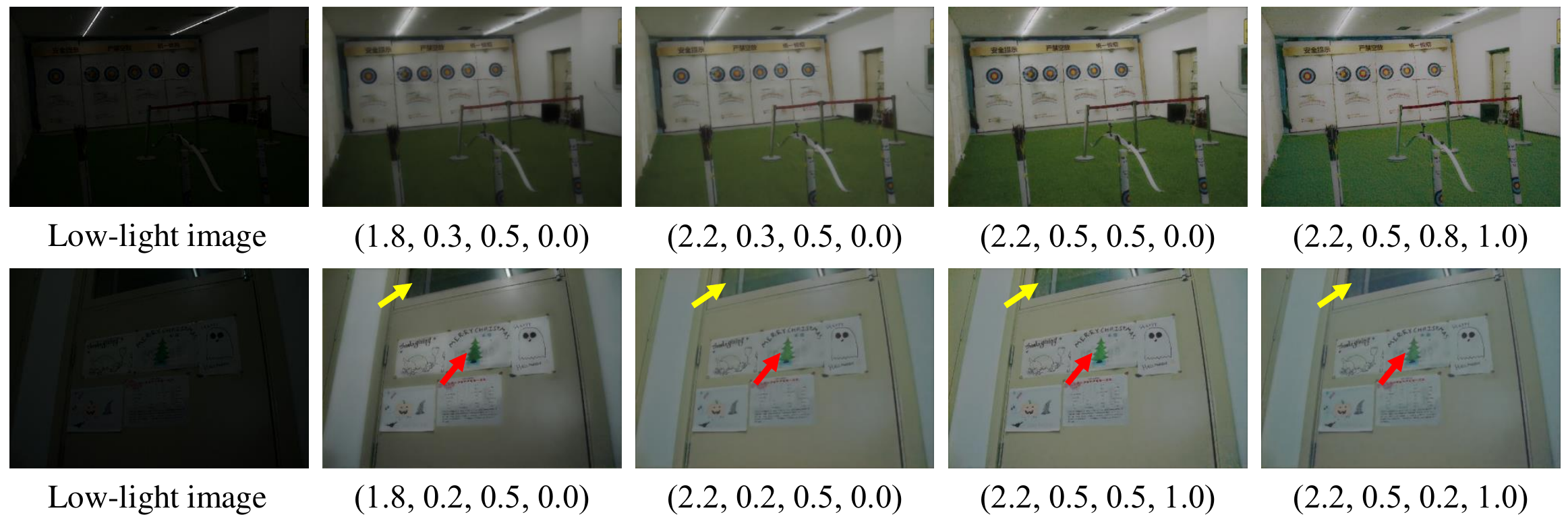}
\vspace{-0.4cm}
\caption{Examples of customized low-light image enhancement by adjusting parameters. The numbers in the parentheses represent brightness, hue, saturation, and noise level, \emph{i.e.}, $(\gamma, C_{h}, C_{s}, C_{n})$.}
\label{fig:per2}
\vspace{-0.3cm}
\end{figure*}

\subsection{Quantitative Evaluation}
To compare our approach with the existing methods quantitatively, we employ two metrics: the peak signal to noise ratio (PSNR, dB) and the structural similarity (SSIM). 
The evaluation of the LOL, FiveK, and Huawei datasets is illustrated in Table~\ref{tab:LOLFiveK}.
It can be seen that the proposed method achieves the best performance with 19.5893 db and 0.7245 SSIM, 18.0260 dB and 0.7262 SSIM, 17.3729 db and 0.5771 SSIM on the above three datasets, which exceed the second-best method with a significant margin. 
It demonstrates that our iUP-Enhancer possesses the highest capability among all unsupervised algorithms and personalized approaches, despite the fact that it does not require any paired training data.

\subsection{Qualitative Evaluation}
Fig.~\ref{fig:LOL} compares subjective visual quality on the challenging extremely low-light images from the LOL dataset. 
It is evident that CRM, SRIE, and RetinexDIP generate the results with underexposure, while ZERO-DCE yields overexposed enhanced results. 
Undeniably, RUAS, cGAN, and ReLLIE predict the well-exposed images. However, the color and contrast deviation problems in their results are non-ignorable. 
Meanwhile, the algorithms cannot remove the noise in the dark well. 
In contrast, our iUP-Enhancer produces the enhanced results with improved brightness, whose hue and saturation are also closer to the ground truth. 
More visual comparison results are shown in the supplementary material.

We conduct user studies to compare the proposed iUP-Enhancer and the baselines. 
A total of 50 human subjects were invited to independently score the personalization of the enhanced results based on their preferred reference images. 
Specifically, for each low-light image, each user selects an unpaired reference image from the reference set for personalization according to his/her aesthetic and repeats this process five times. 
For every testing image, each participant is presented with their customized five enhanced images and the results produced by the baselines , and is then asked to vote for the algorithms yielding the pleasing results. 
Table~\ref{tab:LOLFiveK} summarizes the voting results, which indicates our iUP-Enhancer gets the most votes benefiting from the customization of participants’ aesthetics.

\begin{figure*}[t]
\centering
\includegraphics[width=0.9\textwidth]{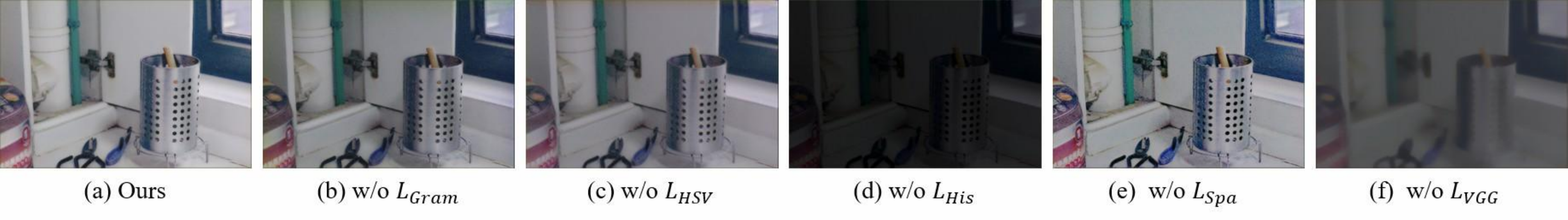}
\vspace{-0.4cm}
\caption{Visualization of the contribution of each loss (histogram loss $L_{his}$, gram matrix loss $L_{gram}$, chromaticity loss $L_{chr}$, spatial consistency loss $L_{spa}$, and perceptual loss $L_{per}$). Please zoom in for better visualization.}
\label{fig:ablation}
\vspace{-0.3cm}
\end{figure*}

\subsection{Visualization of Personalized LLIE}
Figs.~\ref{fig:per1}, \ref{fig:per3}, and \ref{fig:per2} show the personalized results by a single reference image, cross-attribution references, and adjusting parameters.

\textbf{Selecting preferences. } 
Given a low-light image, we achieve personalization in the single-/multiple-reference or cross-attribution reference manner by randomly sampling reference images. 
As illustrated in Fig.~\ref{fig:per1}, the reference images have an effect on the brightness and chromaticity of the results (see the blue and red boxes). 
The bedside and quilt are saturated differently according to the style of the reference images. 
Fig.~\ref{fig:per3} illustrates the personalized result by cross-attribution references. 
The histograms in the second row demonstrate that the brightness and chromaticity of the enhanced result are approximate to the style of their corresponding reference images. 
Meanwhile, the noise level of the enhanced result is reduced.

\textbf{Adjusting parameters. }
Our iUP-Enhancer enables users to customize the enhanced results by configuring parameters for brightness, hue, saturation, and noise level. 
The second and third columns in Fig.~\ref{fig:per2} demonstrate the effectiveness of brightness adjustment. 
The chromaticity of the results can be edited by adjusting hue and saturation parameters. 
As the parameters are increased, the saturation of the windows and trees (yellow and red arrows) increases.
By adjusting the noise-level parameter to 0, the noise level will be greatly reduced (see the yellow arrows).

\begin{figure}[t]
\centering
\includegraphics[width=0.39\textwidth]{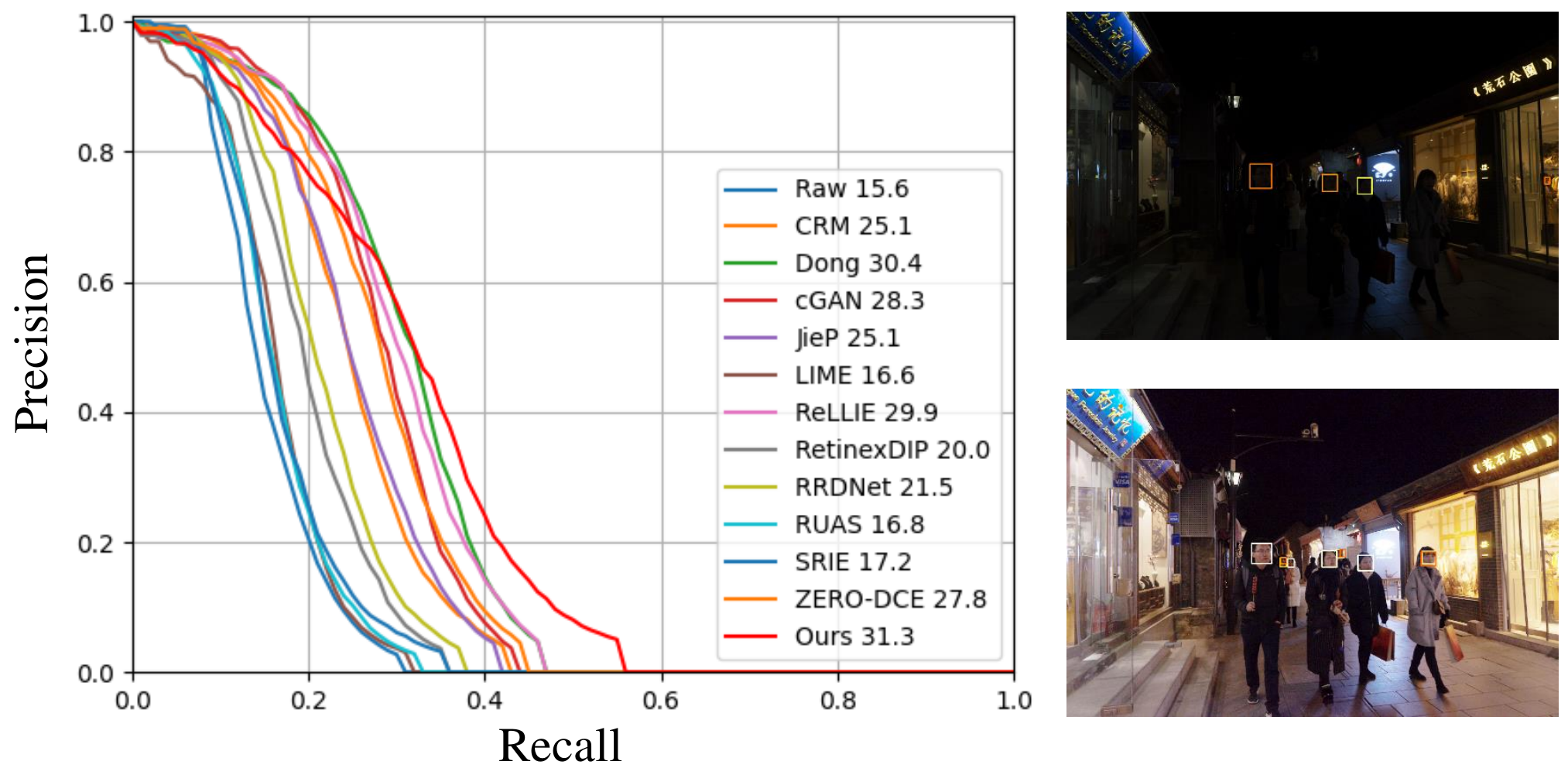}
\vspace{-0.4cm}
\caption{The performance of face detection in the dark. The left image shows the P-R curves and the AP performance. The right two images illustrate the examples of detection before and after enhancement by our iUP-Enhancer.}
\label{fig:darkface}
\vspace{-0.4cm}
\end{figure}

\subsection{Pre-Processing for Improving Face Detection}
Due to the observation that the performance of the high-level tasks falls significantly in low-light conditions, we investigate the impact of our method for face detection on the DARK FACE dataset~\cite{darkFace}. 
The DARK FACE dataset comprises 10,000 images captured in the dark, including 6000 images for training/validation and 4000 images for testing. 
Since the bounding boxes of the testing set are not publicly available, we perform the comparison on the 500 images randomly sampled from the training and validation sets. 
We apply the compared LLIE method as a pre-processing step, and a state-of-the-art face detector, DSFD~\cite{DSFD}, trained on WIDER FACE dataset~\cite{widerface}, as the baseline model. 
The reference image of our method is randomly sampled from the normal-light images in the LOL dataset. 
Fig.~\ref{fig:darkface} depicts the precision-recall (P-R) curves and the average precision (AP) of each algorithm. 
Using our iUP-Enhancer as the pre-processing improves the AP from 15.6$\%$ (DSFD + raw) to 31.3$\%$ (DSFD + iUP-Enhancer), demonstrating that iUP-Enhancer can improve the performance of high-level vision tasks. 
As illustrated in examples, our iUP-Enhancer brightens the extremely low-light regions, hence largely improving the performance of the face detector in the dark. 

\begin{table}[!t]
\caption{Ablation studies of the contribution of loss functions.}
\vspace{-0.3cm}
\begin{center}
\begin{center}
	\setlength{\tabcolsep}{5mm}{
		\begin{tabular}{ccc}
			\toprule 
			Loss Function & PSNR$\uparrow$ & SSIM$\uparrow$  \\
			\midrule
			w/o $L_{His}$				& 8.0780 	& 0.2281 \\
			w/o $L_{Spa}$            	& 17.5836	& 0.5468 \\
			w/o $L_{HSV}$            	& 17.3009 	& 0.6532 \\
			w/o $L_{VGG}$          		& 12.8729 	& 0.5381 \\
			w/o $L_{Gram}$             	& 17.6440 	& 0.6614 \\
			Full Loss          			& 19.5893 	& 0.7245 \\
			\bottomrule
	\end{tabular}}
\end{center}
\end{center}
\vspace{-0.4cm}
\label{tab:ablation}
\end{table}

\subsection{Ablation Studies}

We present the results of our network on the LOL dataset by various combinations of the loss functions in Fig.~\ref{fig:ablation}.  
Removing the illumination histogram loss fails to light up the image, which proves its effectiveness in improving the brightness of the enhanced image. 
The enhanced images without the spatial consistency loss have more noise than the full results. 
This indicates its importance in maintaining the difference of neighboring regions between the input and the enhanced images. 
The hue and saturation of the results cannot be adjusted by the reference images when the HSV histogram loss is discarded. 
Removing the Gram metric loss will degrade the chromaticity correlation between the enhanced and the corresponding reference images, and lead to color distortion. 
The color can be modulated according to the reference images when combining the HSV histogram and Gram metric loss. 
Finally, removing the perceptual loss hampers the structure and detailed information of the input images. 
The quantitative evaluation in Table~\ref{tab:ablation} demonstrates that ignoring any loss functions will result in performance degradation. 
Ablation studies on the multiple computations of correlation and the effectiveness of the BCB, CTB, and IDB are reported in the supplement.

\section{Conclusion}
In this paper, we develop an intelligible unsupervised personalized enhancer (iUP-Enhancer) that enables users to customize the enhanced results for low-light images. 
The iUP-Enhancer extracts the correlations between the input and reference images in brightness, chromaticity, and noise level. 
The iUP-Enhancer trained with the guidance of these correlations presents an intelligible enhancement process and generates customized results varying with the user's aesthetic. 
Notably, our iUP-Enhancer embraces excellent flexibility and scalability by performing personalization with single or multiple reference images, cross-attribution references, or merely adjusting parameters. 
Extensive qualitative and quantitative experiments have validated the superiority of our iUP-Enhancer. 

\begin{acks}
This work was supported by the Anhui Provincial Natural Science Foundation under Grant 2108085UD12. We acknowledge the support of GPU cluster built by MCC Lab of Information Science and Technology Institution, USTC.
\end{acks}

\bibliographystyle{ACM-Reference-Format}
\bibliography{reference}
\end{document}